\documentclass[fleqn,10pt,pdftex,cmex10,table]{wlscirep}
\usepackage[utf8]{inputenc}
\usepackage[T1]{fontenc}
\usepackage{xcolor,soul,framed} 
\colorlet{shadecolor}{yellow}
\DeclareGraphicsExtensions{.pdf,.jpeg,.png}

\usepackage{array}
\usepackage{mdwmath}
\usepackage{eqparbox}
\usepackage{url}
\usepackage{diagbox}
\usepackage{tabularx}
\usepackage{graphicx}
\usepackage{adjustbox}
\usepackage{rotating}
\usepackage{multirow}
\usepackage{subcaption}  
\usepackage{algorithm}
\usepackage{algpseudocode}
\usepackage{bm} 
\usepackage{amsmath} 
\usepackage{booktabs} 

\usepackage{amssymb}

\usepackage{tikz}

\def\checkmark{\tikz\fill[scale=0.4](0,.35) -- (.25,0) -- (1,.7) -- (.25,.15) -- cycle;}

\newcommand{\etal}{et al.}

\usepackage{xcolor}
\usepackage{amssymb}
\usepackage{lscape}

    \title{Tailoring Adversarial Attacks on Deep Neural Networks for Targeted Class Manipulation Using DeepFool Algorithm}
    
\iftrue

\author[1,*]{S. M. Fazle Rabby Labib}
\author[2,*]{Joyanta Jyoti Mondal}
\author[3]{Meem Arafat Manab}
\author[4]{Xi Xiao}
\author[5]{Sarfaraz Newaz}

\affil[1]{School of Data and Sciences, BRAC University, Dhaka, Bangladesh}
\affil[2]{Department of Computer and Information Sciences, University of Delaware, United States}
\affil[3]{School of Law and Government, Dublin City University, Dublin, Ireland}
\affil[4]{Department of Computer Science, College of Arts and Sciences, University of Alabama at Birmingham, United States}
\affil[5]{Next-Generation Computing (NeC) Research Group, Department of Computer Science and Engineering, Bangladesh University of Engineering and Technology, Dhaka, Bangladesh}

\affil[1]{s.m.fazle.rabby.labib@g.bracu.ac.bd}
\affil[2]{joyanta@udel.edu}
\affil[3]{meem.arafat@bracu.ac.bd}
\affil[4]{xxiao@uab.edu}
\affil[5]{DreamZViewerS@gmail.com}

\affil[*]{Corresponding Author}

\fi

\begin{abstract}
The susceptibility of deep neural networks (DNNs) to adversarial attacks undermines their reliability across numerous applications, underscoring the necessity for an in-depth exploration of these vulnerabilities and the formulation of robust defense strategies. The DeepFool algorithm by Moosavi-Dezfooli et al. (2016) represents a pivotal step in identifying minimal perturbations required to induce misclassification of input images. Nonetheless, its generic methodology falls short in scenarios necessitating targeted interventions. Additionally, previous research studies have predominantly concentrated on the success rate of attacks without adequately addressing the consequential distortion of images, the maintenance of image quality, or the confidence threshold required for misclassification. To bridge these gaps, we introduce the Enhanced Targeted DeepFool (ET DeepFool) algorithm, an evolution of DeepFool that not only facilitates the specification of desired misclassification targets but also incorporates a configurable minimum confidence score. Our empirical investigations demonstrate the superiority of this refined approach in maintaining the integrity of images and minimizing perturbations across a variety of DNN architectures. Unlike previous iterations, such as the Targeted DeepFool by Gajjar et al. (2022), our method grants unparalleled control over the perturbation process, enabling precise manipulation of model responses. Preliminary outcomes reveal that certain models, including AlexNet and the advanced Vision Transformer, display commendable robustness to such manipulations. This discovery of varying levels of model robustness, as unveiled through our confidence level adjustments, could have far-reaching implications for the field of image recognition. Our code is available at \url{https://github.com/FazleLabib/et_deepfool}.

\end{abstract}

\keywords{Adversarial Attack, Image Classification, Deep Neural Network}

\begin{document}
\flushbottom
\maketitle
%
%
\thispagestyle{empty}

\section{Introduction}
\label{sec:intro}

    \begin{figure*}[ht]
        \centering
        \includegraphics[width=\linewidth]{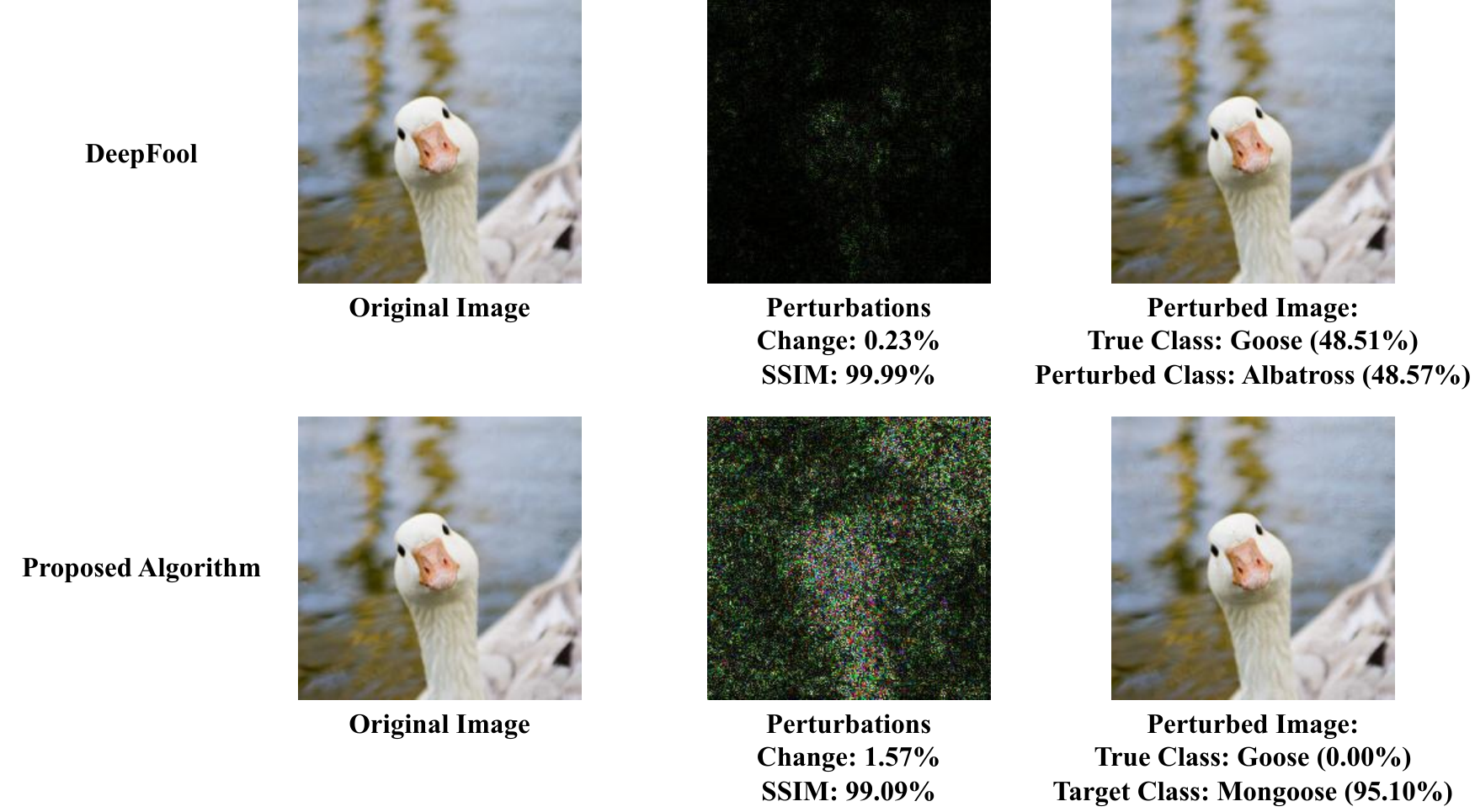}
        \caption{Comparison between original DeepFool and our proposed Enhanced Targeted DeepFool. Here, the sample image is taken from the ImageNet dataset and the perturbation image is scaled 20 times for visibility}
        \label{fig:DeepFool_vs_TargetedDeepFool}
    \end{figure*}

Deep neural networks (DNNs) have revolutionized many fields including but not limited to speech recognition~\cite{hinton2012, Chan2016}, computer vision~\cite{chai2021, Hashmi2022-ed}, natural language processing~\cite{vaswani2017}, and even game playing~\cite{silver2017mastering}. However, their high accuracy and robustness can be compromised by adversaries who intentionally manipulate the input data to fool the model. Such attacks can have serious consequences in real-world applications such as autonomous driving, medical diagnosis, and security systems. Therefore, understanding the vulnerabilities of DNNs to adversarial attacks and developing effective defense mechanisms has become an important research area in machine learning and computer security. DeepFool is one of the algorithms, proposed by Moosavi-Dezfooli \etal~\cite{deepfool}, which iteratively finds the minimum amount of perturbations required to push a given input image to a misclassified region of the feature space. They use the following equation that defines an adversarial perturbation as the minimal perturbation \textbf{$r$} that is sufficient to change the estimated label \textbf{$\hat{k}(x)$}:
    \begin{equation}
        \Delta(x;\hat{k}) := \min_{r} ||r||_2 \textrm{ subject to } \hat{k}(x+r) \neq \hat{k}(x) \quad
    \label{eq:perturbation_minimization}
    \end{equation}
where, $x$ is an image, and \textbf{$\hat{k}(x)$} is the estimated label. With this, an image can be misclassified with a minimal amount of perturbations. However, this approach is not focused on any specific target. Instead, the images are classified as a different class with a minimal amount of perturbation. Thus, if an image $x$ can be misclassified as some class \textit{A} with less perturbation than some other class \textit{B}, DeepFool will choose to use the perturbation that misclassifies $x$ as class \textit{A}. 



In such an evolving landscape of artificial intelligence and its applications\cite{Ahmed2021-fo,Siddique2022-qv,Rahman2023-fn,Zoana2023-wn,Hossain2018-ih,Mondal2024-pr,Mondal2022-cv,Ul_Islam2022-hd,10.1145/3530190.3534844}, the vulnerability of deep neural networks (DNNs) to adversarial attacks has emerged as a significant concern. While untargeted attacks introduce small perturbations to induce misclassification, targeted attacks pose a more nuanced threat by aiming to trick the DNN into categorizing input data as a specific, incorrect label. This specificity increases the potential damage, such as misleading autonomous vehicles into misinterpreting traffic signs/license plates \cite{Nasim_2024_WACV} or circumventing facial recognition-based security systems. Thus, the development of precise and controlled methods for executing targeted attacks is imperative for evaluating and enhancing the resilience of DNNs against such manipulations.

Although the DeepFool algorithm represents a step forward in identifying minimal perturbations necessary for general misclassification, its framework lacks the specificity required for targeted attacks. The algorithm introduced by Gajjar et al.,~\cite{tardeep}, named Targeted DeepFool marks an advancement in this domain by enabling the misclassification of images into predetermined categories. Despite this progress, the algorithm's practical application is hindered by a relatively low success rate and a static nature that precludes adjustment of its internal hyperparameters. These limitations highlight the need for a more flexible and effective solution.

To bridge the existing gap, this manuscript introduces the Enhanced Targeted DeepFool (ET DeepFool) method, an innovative variant of the DeepFool algorithm. This variant not only allows targeting a specific misclassification class but also enhances DeepFool by incorporating a parameterization feature that permits the setting of minimum confidence score thresholds. We illustrate that ET DeepFool is more straightforward than its predecessor in terms of computational complexity and more adept at inducing misclassifications across various deep neural network architectures towards chosen classes. Subsequently, the efficacy of the method is evaluated. Our empirical evidence suggests that ET DeepFool operates with high efficiency across diverse computational platforms while maintaining the visual integrity of images close to their original state. It introduces a novel approach for assessing model robustness through perturbation, marking the first instance, to our knowledge, of a perturbation technique that allows for the explicit specification of performance metrics such as confidence scores. This innovation enables the determination of a perturbation magnitude sufficient to deceive an image classification model into misclassifying an image into a different class with significant error rates and confidence scores. Unlike previous methods that merely focused on the act of fooling the model often resulting in low-confidence misclassifications, our method ensures high confidence in the misclassifications it generates. The contributions of this work are summarized as follows:

\begin{itemize}
\item Introduction of the Enhanced Targeted DeepFool algorithm, a sophisticated targeted adversarial attack method that evolves from the original DeepFool algorithm. This development not only extends the functionality of DeepFool but also introduces a novel approach to targeted attacks in deep learning models.

\item Enhancement of the algorithm through the introduction of a feature for setting minimum confidence score thresholds. This allows for more precise control over the perturbation process, ensuring that the adversarial examples generated are both effective and realistic, with a higher likelihood of fooling the target model while maintaining a degree of stealthiness in the modifications made.

\item Validation of our method's superior performance through comprehensive experiments conducted with eleven image classification models across six distinct datasets. These tests demonstrate not only the versatility of ET DeepFool but also its effectiveness in a wide range of scenarios, highlighting its potential as a valuable tool for security researchers and practitioners in evaluating the robustness of neural networks.

\end{itemize}

The remainder of the paper is divided into the following sections: An overview of pertinent studies carried out in this field is provided in Section~\ref{Related_Work}. Then, in Section~\ref{background}, we go through the context of DeepFool architecture. 
In Section~\ref{Methodology}, we outline our proposed approach, ET DeepFool, which employs the utilization of the DeepFool algorithm to create a targeted adversarial attack that is enhanced with the ability to tune the minimum confidence threshold.
We describe our experimental setup and experimental results in Section~\ref{experiment} and Section~\ref{sec:results}.
Subsequently, we delve deeper into our research findings in Section~\ref{discussion}. Finally, we explore potential future implications in Section~\ref{Conclusion} before concluding the paper.

    \begin{figure*}[!t]
        \centering
        \includegraphics[width=\linewidth]{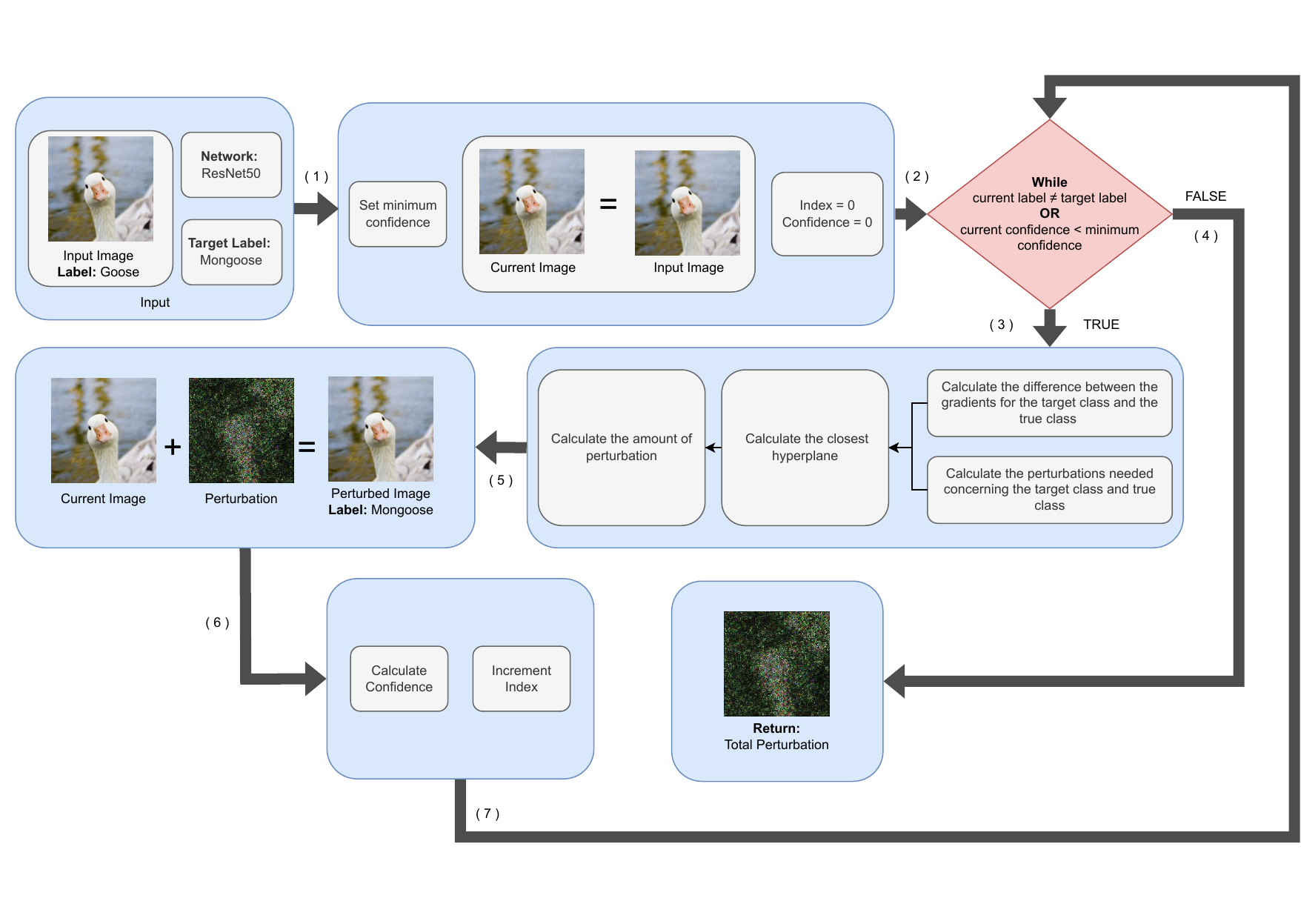}
        \caption{High-level overview of the Enhanced Targeted DeepFool algorithm. The sample image is taken from the ImageNet dataset.}
        \label{fig:ETDeepFool_flowchart}
    \end{figure*}
    
\section{Related Work}
\label{Related_Work}

Adversarial attacks involve perturbing data to some extent to cause misclassification by an ML model. These attacks can be implemented in several ways in the form of black-box, white-box, and grey-box attacks. Furthermore, there are data poisoning attacks which include label flipping, clean label, and backdoor attacks. 
In this section, we cover existing literature related to different adversarial attacks against image classification models as well as works done on adversarial defense. 


\subsection{White-Box Attacks}
In white-box attacks, the attackers have complete knowledge of the target model’s architecture, weights, gradients, parameters, and training data. By having access to the model's internals, an attacker can explore its vulnerabilities more effectively and create adversarial examples that are highly effective at deceiving the model's predictions. There are several common white-box adversarial attack methods used in the field of image classification. One such method is the Fast Gradient Sign Method (FGSM), a type of adversarial attack for image classification that involves adding minimal noise to each pixel of an image, based on the gradient of the loss function concerning the image~\cite{goodfellow}. 
Another notable algorithm proposed by Carlini and Wagner finds the smallest noise to be added to an image to misclassify it~\cite{carlini2017}. This method goes beyond the FGSM approach by seeking the most effective perturbation for achieving misclassification. Jacobian-based Saliency Map Approach as proposed by Papernot et al., works by iteratively modifying the input features of a sample to maximize the difference between the predicted output and the true output~\cite{jsma}. 
Additionally, the Universal Adversarial Perturbations by Moosavi-Dezfooli et al., fool a deep neural network by adding the same perturbations to multiple images, causing it to misclassify all of the affected images~\cite{moosavidezfooli2017universal}. Furthermore, Duan et al., propose an attack that drops information from the image instead of perturbing it~\cite{advdrop}.

\subsection{Black-Box Attacks}
In black box attack scenarios, the internal workings of the models are not available. The attacker usually has the input-output behavior and the probability labels of the target models. Gao et al., present a black-box attack method called Patch-wise Iterative Fast Gradient Sign Method that outperforms pixel-wise methods in generating transferable adversarial examples against various mainstream models~\cite{patchwise}. 
Another approach by Zhao et al., is a GAN-based method that involves training a generator network to produce perturbations that can be added to the original input to create an adversarial example~\cite{zhao18}. This method proposed by Li \etal, integrates Poincaré distance into iterative FGSM and uses a metric learning approach to regularize iterative attacks~\cite{PoTrip}. It generates transferable targeted adversarial examples by iteratively perturbing the input image in the direction of the target class while simultaneously minimizing the Poincaré distance between the original and perturbed images. The Adversarial Patch attack, as proposed by Brown \etal, takes a different approach~\cite{Brown2018}. Instead of modifying the entire image, it focuses on creating a small patch that can be strategically placed in the real world. 
Furthermore, Su et al., create a method that fools a network by only altering a single pixel of an image~\cite{Su_2019}.
Wei et al., propose a very different approach by manipulating image attributes such as brightness, contrast, and sharpness instead of generating any adversarial noise~\cite{weimani}.

\subsection{Data Poisoning Attacks}

Shafahi et al.,~\cite{poisonfrog} apply a one-shot poisoning attack by injecting a single poison instance into a clean dataset, causing the model to misclassify a specific target instance without negatively impacting its performance on other examples. 
Huang et al., propose MetaPoison, a meta-learning approach for crafting poisons to fool neural networks using clean-label data poisoning~\cite{metapoison}. 
In~\cite{hiddenpoison}, Di \etal, presents a camouflaging approach for targeted poisoning attacks based on the gradient-matching approach of Geiping \etal,~\cite{geiping}. 
Muñoz-González et al., propose pGAN, a scheme that generates poisoning points to maximize the error of a target classifier while minimizing detectability by a discriminator~\cite {munoz}. 

\subsection{Adversarial Defense}
To protect DNNs from adversarial attacks and improve their robustness various methods are applied. These methods include training a network with adversarial examples, detecting adversarial examples instead of classifying them, and using randomness to defend the networks. One notable improvement in performance was introduced by Ding et al., which combines the cross-entropy loss with a margin maximization loss term, which is applied to correctly classified examples~\cite{mma}. In a different vein, Xu et al., propose a method called feature squeezing, which decreases the search space of an adversary by combining samples analogous with multiple different vectors into a single sample~\cite{featureSqueeze}. Furthermore, Zheng et al., propose a method that requires modification of the output of the classifier by performing hypothesis testing and using Gaussian mixture models to detect adversarial examples~\cite{zheng}. \\

\noindent ET DeepFool builds upon these foundational concepts by extending the principles of white-box attacks to create a more targeted and efficient adversarial method. While previous works, such as FGSM and Carlini-Wagner, focus on minimizing perturbations to deceive models, our method advances this approach by simplifying the attack process, reducing computational complexity, and ensuring high-confidence misclassification.

\section{Background and Motivation}
\label{background}

In this section, we initially discuss the background of the original DeepFool method which we mention as the Vanilla DeepFool, and the Targeted DeepFool method proposed by Gajjar \etal ~\cite{tardeep}. 

\subsection{Vanilla DeepFool}
    
    In a multi-class classification setting as seen on Algorithm~\ref{algorithm1}, each class can be thought of as being separated by a boundary, known as a hyperplane. When an input, represented as $\bm{x}$, is classified, it gets assigned to the class on the side of the boundary where it falls. The DeepFool algorithm works by finding the nearest boundary and slightly adjusting $\bm{x}$ so that it crosses this boundary and gets misclassified. This process is repeated until the input is misclassified, and the algorithm returns the total amount of changes (called perturbations), denoted as $\bm{\hat{r}}$, needed to cause the misclassification.
    To find the closest boundary, denoted as $\hat{l}$, the algorithm first calculates a vector $\bm{w}'_k$ that points toward the boundary between the current predicted class and another class $k$. This vector is found by subtracting the gradient of the current predicted class from the gradient of class $k$:
    \begin{equation}
        \bm{w}'_k \leftarrow \nabla f_k(\bm{x}_i) - \nabla f_{\hat{k}(\bm{x}_0)}(\bm{x}_i)
    \label{eq:weight_update_gradient_difference}
    \end{equation}
    Here, $\nabla f_k(\bm{x}i)$ is the gradient of the activation for class $k$, and $\nabla f{\hat{k}(\bm{x}_0)}(\bm{x}_i)$ is the gradient of the activation for the currently predicted class.
    Next, the algorithm calculates the difference between the activation values of the two classes:
    \begin{equation}
        f'_k \leftarrow f_k(\bm{x}_i) - f_{\hat{k}(\bm{x}_0)}(\bm{x}_i)
    \label{eq:activation_difference}
    \end{equation}
    After obtaining $\bm{w}'_k$ and $f'k$, the algorithm determines the closest boundary $\hat{l}$ and the smallest change needed in this direction for the $k{th}$ iteration:
    \begin{equation}
        \hat{l} \leftarrow \operatorname*{arg\,min}_{k\neq\hat{k}(\bm{x}_0)}\frac{|f'_k|}{||\bm{w}'_k||_2}
    \label{eq:target_class_selection}
    \end{equation}
    \begin{equation}
        \bm{r}_i \leftarrow \frac{|f'_{\hat{l}}|}{||\bm{w'}_{\hat{l}}||_2^2}\bm{w'}_{\hat{l}}
    \label{eq:perturbation_vector_computation}
    \end{equation}
    This process continues until the predicted class $\hat{k}(\bm{x}_i)$ changes to a different label, at which point the algorithm stops and returns the total perturbation $\bm{\hat{r}}$.
    
\subsection{Targeted DeepFool}

Gajjar et al.~\cite{tardeep} propose two algorithms for generating adversarial examples. These are known as the basic iterative approach and the recursive approach.

\noindent\textbf{Basic Targeted DeepFool Algorithm:} This is an iterative approach as seen on Algorithm~\ref{algorithm2}, which works by making small changes or perturbations to an image. The goal of these perturbations is to gradually alter the image so that a classifier starts to see it as belonging to a different, specific target class rather than its original class. The process is iterative: in each step, the algorithm computes the direction and magnitude of the perturbation that will push the image closer to being misclassified as the target class. This is done by moving the image towards the hyperplane of the classifier that separates the original class from the target class. The algorithm continues this process, until one of two things happens: either the classifier incorrectly labels the image as the target class, or the amount of perturbation applied to the image reaches a pre-set limit, beyond which further changes might make the image too distorted.

\noindent\textbf{Recursive Targeted DeepFool Algorithm:} This approach extends the basic idea by addressing situations where the basic algorithm might get ``stuck'' on an intermediate step, unable to push the image all the way to the target class. This can happen because the decision boundaries in a deep neural network are often non-linear, making it difficult for a simple iterative process to find a path to the target class. In the recursive approach, if the basic algorithm fails to reach the target classification after a certain number of iterations, the algorithm resets and starts the process again, but this time from the current altered state of the image. This refines the image's perturbations in each pass, which helps the image navigate around difficult regions of the decision boundaries. This process continues until the image is successfully misclassified as the target class or a maximum number of recursion attempts is reached. This recursive strategy makes the approach more robust and often more effective than the basic iterative method.

 


\section{Methodology}
\label{Methodology}

In this section, we introduce our proposed Enhanced Targeted DeepFool.

\subsection{Enhanced Targeted DeepFool}
    To adapt the original DeepFool algorithm for misclassifying an image into a specific target class, we propose the following modified version of the algorithm, detailed in Algorithm~\ref{algorithm3}. A high-level overview of the algorithm is shown in figure~\ref{fig:ETDeepFool_flowchart}.

    \begin{figure*}
    \begin{minipage}{0.32\textwidth}
    \begin{algorithm}[H]
    \centering
    \caption{\\\hspace*{0.5cm}DeepFool: multi-class case}\label{algorithm1}
            \begin{algorithmic}[1] 
                \State \textbf{Input:} Image $\bm{x}$, classifier $f$
                \State
                \State \textbf{Output:} Perturbation $\bm{\hat{r}}$
                \State Initialize $\bm{x}_0 \leftarrow \bm{x}, i \leftarrow 0$
                \While{$\hat{k}(\bm{x}_i) = \hat{k}(\bm{x}_0)$} 
                \For{$k \neq \hat{k}(\bm{x}_0)$}
                \State $\bm{w}'_k \leftarrow \nabla f_k(\bm{x}_i) - \nabla f_{\hat{k}(\bm{x}_0)}(\bm{x}_i)$
                \State $f'_k \leftarrow f_k(\bm{x}_i) - f_{\hat{k}(\bm{x}_0)}(\bm{x}_i)$
                \EndFor
                \State $\hat{l} \leftarrow \operatorname*{arg\,min}_{k\neq\hat{k}(\bm{x}_0)}\frac{|f'_k|}{||\bm{w}'_k||_2}$
                \State $\bm{r}_i \leftarrow \frac{|f'_{\hat{l}}|}{||\bm{w'}_{\hat{l}}||_2^2}\bm{w'}_{\hat{l}}$
                \State $\bm{x}_{i+1} \leftarrow \bm{x}_i + \bm{r}_i$
                \State  $i \leftarrow i + 1$
                \EndWhile
                \State \textbf{return} $\bm{\hat{r}} = \sum_i \bm{r}_i$
            \end{algorithmic}
    \end{algorithm}
    \end{minipage}\hspace*{0.2cm}
    \begin{minipage}{0.36\textwidth}
    \begin{algorithm}[H]
    \caption{\\\hspace*{0.5cm}Basic Targeted DeepFool}\label{algorithm2}
    \begin{algorithmic}[1]
    \State \textbf{Input:} Image $\bm{x}$, classifier $h$, 
    \State \hspace*{0.5cm}target class $t$, total perturbation $p_{tot}$
    \State \textbf{Output:} Total perturbation $p_{tot}$
    \State Initialize $\bm{x}_0 \leftarrow \bm{x}$, $i \leftarrow 0$
    \State Set threshold (maximum perturbation)
    \While{$\arg\max_j h_j(\bm{x}_i) \neq t$ 
        \\\hspace*{2cm} or $p_i < \text{threshold}$}
        \State
        \State $p_i \leftarrow \frac{|h_t(\bm{x}_i) - h_j(\bm{x}_i)({x}_i)|}{||g_t - g_j(\bm{x}_i)||_2^2} (g_t - g_j(\bm{x}_i))$
        \State
        \State $\bm{x}_{i+1} \leftarrow \bm{x}_i + p_i$
        \State $i \leftarrow i + 1$
    \EndWhile
    \State $p_{tot} \leftarrow \sum_i p_i$
    \State
    \State \textbf{return} $p_{tot}$
    \end{algorithmic}
    \end{algorithm}
    \end{minipage}\hspace*{0.2cm}
    \vspace*{0.2cm}
    \begin{minipage}{0.32\textwidth}
    \begin{algorithm}[H]
    \centering
    \caption{Proposed: \\\hspace*{0.5cm}Enhanced Targeted DeepFool}\label{algorithm3}
            \begin{algorithmic}[1] 
                \State \textbf{Input:} Image $\bm{x}$, classifier $f$ 
                \State \hspace*{0.2cm}target class $t$, min confidence $\bm{c}_{min}$
                \State \textbf{Output:} Perturbation $\bm{\hat{r}}$
                \State Initialize $\bm{x}_0 \leftarrow \bm{x}, i \leftarrow 0, c \leftarrow 0.$ 
                \While{$\hat{k}(\bm{x}_i) \neq t$ or $c < \bm{c}_{min}$} 
                \State $\bm{w}'k \leftarrow \nabla f_{t}(\bm{x}_i) - \nabla f_{k(\bm{x}_0)}(\bm{x}_i)$
                \State $f'_k \leftarrow f_{t}(\bm{x}_i) - f_{k(\bm{x}_0)}(\bm{x}_i)$
                \State
                \State $\hat{l} \leftarrow \frac{|f'_k|}{||\bm{w}'_k||_2}$
                \State $\bm{r}_i \leftarrow \frac{|f'_{\hat{l}}|}{||\bm{w'}_{\hat{l}}||_2^2}\bm{w'}_{\hat{l}}$
                \State $\bm{x}_{i+1} \leftarrow \bm{x}_i + \bm{r}_i$
                \State $c \leftarrow softmax(0, t)$
                \State  $i \leftarrow i + 1$
                \EndWhile
                \State \textbf{return} $\bm{\hat{r}} = \sum_i \bm{r}_i$
            \end{algorithmic}
    \end{algorithm}
    \end{minipage}    
    \\
    \centering Comparison among the original DeepFool, basic Targeted DeepFool, and our proposed Enhanced Targeted DeepFool algorithms
    \label{fig:algorithms}
    \end{figure*}
        
    To start with, we run the \texttt{while} loop until the current label is not equal to the target label instead of running it until the image is misclassified. Additionally, we have eliminated the inner \texttt{for} loop, as compared to line 6 of Algorithm~\ref{algorithm1}. This loop calculates the gradients for all classes except the original one, which is not necessary in the proposed case when we focus on a single target class. In Algorithm~\ref{algorithm1} there is a \texttt{for} loop inside a \texttt{while} loop. The \texttt{while} loop is responsible for the complexity as $O(N)$ and the inner \texttt{for} loop is responsible for the complexity as $O(C)$, both create the complexity as $O(NC)$ for Algorithm~\ref{algorithm1}. Here $N$ is the maximum number of iterations and $C$ is the number of classes. On the other hand, in our proposed method (Algorithm~\ref{algorithm3}), we remove the inner \texttt{for} loop from the inside of the \texttt{while} loop as compared to Algorithm~\ref{algorithm1}. Therefore, the complexity of the proposed method depends only upon the \texttt{while} loop and its $O(N)$. The removal of the inner \texttt{for} loop from inside the \texttt{while} loop significantly reduces the computational complexity of Algorithm~\ref{algorithm3}, effectively decreasing the complexity by a factor of $O(C)$, where $C$ is the number of classes. 
    
    We modify Equations~\ref{eq:weight_update_gradient_difference} and \ref{eq:activation_difference} of the original algorithm as follows to reflect this targeted approach. Here, $\bm{w}'_k$ now computes the difference between the gradients of the target class and the true class, instead of comparing with multiple potential classes. Similarly, $f'_k$ is updated to measure the perturbations needed specifically concerning the target class and the true class.
    \begin{equation}
        \bm{w}'k \leftarrow \nabla f_{t}(\bm{x}_i) - \nabla f_{k(\bm{x}_0)}(\bm{x}_i)
    \label{eq:updated_weight_update_gradient_difference}
    \end{equation}
    \begin{equation}
        f'_k \leftarrow f_{t}(\bm{x}_i) - f_{k(\bm{x}_0)}(\bm{x}_i)
    \label{eq:updated_activation_difference}
    \end{equation}
    Since the algorithm no longer needs to compare across multiple classes, Equation~\ref{eq:target_class_selection} is simplified as follows:
    \begin{equation}
        \hat{l} \leftarrow \frac{|f'_k|}{||\bm{w}'_k||_2}
    \label{eq:updated_target_class_selection}
    \end{equation}
    These adjustments allow us to effectively steer the misclassification of an image towards a specific target class of our choosing with only minimal changes to the original DeepFool algorithm.
    
    Moreover, we have introduced an additional criterion in the algorithm, termed as the minimum confidence level, denoted by $\bm{c}_{min}$. The confidence of the misclassification is determined using the softmax function, where the probability associated with the target class $t$ is calculated:
    \begin{equation}
        c \leftarrow \text{softmax}(0, t)
    \label{eq:softmax_calculation}
    \end{equation}
    This mechanism allows the user to define a minimum confidence threshold as a hyperparameter. The algorithm ensures that the perturbed image not only gets misclassified to the target class but also achieves a confidence level that meets or exceeds the user-defined threshold, thereby producing more robust and reliable adversarial examples.


\section{Experimental Setup}
\label{experiment}

Here we apply our proposed method to multiple state-of-the-art image classification models and show our findings.
    
\subsection{Dataset}

We use six different datasets; MNIST~\cite{mnist}, CIFAR-10~\cite{cifar10}, EuroSAT~\cite{eurosat}, GTSRB~\cite{gtsrb}, HAM10000~\cite{ham10000} and the ILSVRC2012~\cite{ILSVRC15} also known as the ImageNet dataset for our experiments. The MNIST dataset consists of a collection of $28\times28$ grayscale images of handwritten digits (0 to 9), along with their corresponding labels. It contains 60,000 training images and 10,000 test images. The CIFAR-10 dataset is a collection of color images, each $32\times32$ pixels in size, spread across 10 different classes. These classes include common objects such as airplanes, automobiles, birds, cats, deer, dogs, frogs, horses, ships, and trucks with 50000 training images and 10000 test images. The EuroSAT dataset is a benchmark dataset for satellite image classification, derived from the Sentinel-2 satellite imagery. It contains 27,000 labeled $64\times64$ images across 10 classes, including various land cover types like residential areas, forests, and water bodies. The German Traffic Sign Recognition Benchmark (GTSRB) is a dataset used for training and evaluating traffic sign recognition systems. It consists of over 50,000 images of 43 different traffic sign classes, collected under various lighting and weather conditions. The dataset is commonly used for autonomous driving applications. The HAM10000 (Human Against Machine with 10000 training images) dataset comprises 10,015 dermatoscopic images collected from diverse populations and imaging modalities. These images are categorized into various skin lesion types, including melanoma, nevus, and benign keratosis. The ImageNet dataset contains 1,281,167 training images, 50,000 validation images and 100,000 test images with a thousand different classes. It is widely regarded as a benchmark dataset in the field of computer vision and has played a crucial role in advancing the development of deep learning models. The dataset is large and diverse which makes it a comprehensive representation of real-world visual data. Due to its size and diversity, the models pre-trained on this dataset can learn rich feature representation that captures a good amount of visual information. 

For our experiments, we used the test sets provided with MNIST, CIFAR-10, EuroSAT, GTSRB, and HAM10000. On the other hand, as the label of the test set of the ImageNet dataset is not public, therefore, we used the validation dataset for the experiment of our study. It is worth mentioning that, the validation images of the ImageNet dataset are used in several research studies in the literature~\cite{uap, pmlr-v119-shankar20c, Stock_2018_ECCV, NEURIPS2023_863da9d4}.



\subsection{Models} 

For the MNIST dataset, we use a pre-trained LeNet-5~\cite{lenet5} and a CNN model with defensive distillation, achieving 79.14\% accuracy. For the CIFAR-10 dataset, we use pre-trained GoogLeNet~\cite{googlenet} and ResNet-20~\cite{ResNet50} models, as well as a Network in Network~\cite{nin} model achieving 88.51\% accuracy. Additionally, we train a CNN model with defensive distillation \cite{distillation}, achieving 99.17\% accuracy. To experiment with the Enhanced Targeted Deepfool method on the ImageNet dataset, we use pre-trained deep convolutional neural networks, including ResNet-50~\cite{ResNet50}, AlexNet~\cite{alexNet}, EfficientNet-v2~\cite{efficientnetv2}, GoogLeNet~\cite{googlenet}, and Inception-v3~\cite{inceptionV3}. We also test our method using the Vision Transformer (ViT)~\cite{vit} image classification model. For real-world datasets, we use a fine-tuned ResNet-50 model to test our attack on the EuroSAT satellite image dataset and a pre-trained ResNet-50 model on the HAM10000 medical image dataset. Additionally, we train a custom CNN model, as shown in Table~\ref{tab:gtsrb_model_architecture}, achieving 99.05\% accuracy with a learning rate of 0.0008, dropout rates of 0.2 and 0.3, a batch size of 64, and 20 epochs to test our algorithm on the GTSRB autonomous driving dataset.

\begin{table*}[ht]
    \centering
    \begin{tabular*}{0.5\textwidth}{@{\extracolsep{\fill}}ll}
        \toprule
        Layer Type & Layer Configuration \\
        \midrule
        Convolution + ReLU & 3×3×32 \\
        Convolution + ReLU & 3×3×64 \\
        Max Pooling & 2×2 \\
        Convolution + ReLU & 3×3×128 \\
        Convolution + ReLU & 3×3×256 \\
        Max Pooling & 2×2 \\
        Convolution + ReLU & 3×3×512 \\
        Convolution + ReLU & 3×3×1024 \\
        Max Pooling & 2×2 \\
        Fully Connected + ReLU & 512 \\
        Fully Connected + ReLU & 128 \\
        Softmax & 43 \\
        \bottomrule
    \end{tabular*}
    \caption{Custom CNN model architecture  that is used over the GTSRB dataset}
    \label{tab:gtsrb_model_architecture}
\end{table*}

\subsection{Testbed Setups} 
We use four different testbed devices to experiment with our classifiers for this targeted attack. One includes Intel Core i7-12700K processor, RTX 3070 Ti, and 32 GB RAM. Another consists of an Intel Core i5 13400F, RTX 3060 Ti, and 32 GB RAM. The next device includes Ryzen 5, GTX 1050 Ti, and 16 GB RAM. And the last one consists of an Intel CPU with Nvidia Tesla P100 GPU, and 16 GB RAM. We install PyTorch 2.0 and Torchmetrics 0.11.4 libraries in these testbed systems, keeping the version of Python on 3.10.

\subsection{Setting up Hyperparameter and Test Approach} 
For the tests, we use the images and generate a random target class that is not its true class. These images along with the target classes that were generated are fed into our function. We use several hyper-parameters such as overshoot which is set to the default value of 0.02. This hyper-parameter is used as a termination criterion to prevent vanishing updates and oscillations. We set the minimum amount of confidence needed as 95\% and the maximum iterations as 100. This is done because in most cases the confidence score of the perturbed image is usually lower than expected ($\thicksim$ 50\%), therefore we add another condition in the while loop to make the code run until the desired confidence is reached. Although, this will lead to more perturbations. The code will run until these conditions are met or until maximum iterations are reached. These hyper-parameters can be tuned to one's needs.

\noindent\textbf{Guidance on selecting confidence threshold:}
In the case of selecting the confidence threshold, we initially conducted the experiments without setting a confidence threshold as part of an ablation study. Then we choose the confidence threshold as 50\% because, any confidence at least over 50\% is intuitively an acceptable confidence for reliable classification. Further, we increase the confidence threshold by 20\% arbitrarily to assess its impact at the next steps.

\subsection{Metrics} 

There are different approaches to identifying the efficiency of our proposed approach with existing algorithms.


\noindent\textbf{Confidence:}
We calculate the confidence score for the target class by passing the output tensor through the softmax function. It reflects the classifier's level of confidence in its predictions for the perturbed images. 

\noindent\textbf{Perturbation:}
The magnitude of perturbations added to the images referred to as "Perturbations", quantifies the level of changes required to deceive the classifier. We find the change in an image by calculating the L2 distance between the perturbed and the original image and dividing it by the maximum L2 distance. 

\noindent\textbf{Structural Similarity Index Measure (SSIM):}
We calculate the SSIM~\cite{ssim} between the perturbed and original image. This widely used metric in image processing quantifies the similarity between two images by considering three specific aspects: how similar the brightness (luminance) is, how much the local variations in brightness (contrast) are preserved, and how well the underlying structure of the image (e.g., edges, textures) is maintained. By comparing local patterns of pixel intensities rather than relying solely on global characteristics, SSIM provides a more nuanced understanding of image fidelity.

\noindent\textbf{Iterations:}
We record the number of iterations needed to perturb an image. This metric indicates the mean number of iterations required to achieve a successful misclassification. 

\noindent\textbf{Success:}
This metric shows the percentage of images being successfully misclassified as the randomly selected target image. 

\noindent\textbf{Time:}
This metric is used to denote the computational time needed to execute the attack against a single image.

\section{Results}
\label{sec:results}
In this section, we describe the findings after successfully experimenting with our proposed model under different scenarios.

    After running the dataset in different classifiers through our algorithm, we observe several notable results, as presented in Table~\ref{tab:mean_metrics}. Our method generates perturbed images with a mean confidence score of 0.97 for nearly all models.

    We observe that certain classifiers, such as Network in Network (NiN), LeNet-5, ResNet-20, ResNet-50, EfficientNet-v2, GoogLeNet, and Inception-v3, exhibit considerable vulnerability to our approach. The perturbation rates for these models range from 0.01\% to 3.76\%. In contrast, ET Deepfool performs less effectively with AlexNet and ViT, which require perturbation rates of 9.08\% and 11.27\%, respectively, to be fooled. Regarding image integrity, models such as ResNet-20, ResNet-50, EfficientNet-v2, GoogLeNet, and Inception-v3 consistently exhibit high SSIM scores, indicating less noticeable perturbations.

    Focusing on the ImageNet dataset, AlexNet and ViT have the lowest mean SSIM scores, at 92\% and 89\%, respectively. For the MNIST dataset, the CNN with Defensive Distillation \cite{distillation} achieves high confidence (0.99) and low perturbation (1.28\%), however, both the LeNet-5 and CNN (Defensive Distillation) model with MNIST dataset have moderate SSIM scores around 0.63 to 0.64.

    When examining the iteration count required for successful attacks, EfficientNet-v2, GoogLeNet, and Inception-v3 perform consistently well with iteration counts ranging from 33 to 38 for ImageNet and 16 for CIFAR-10. NiN and ResNet-20 on CIFAR-10 require fewer iterations, that is 12 and 14, respectively. ViT stands out with the highest iteration count, averaging 67 iterations per image, indicating more resistance to attacks. In terms of success rate, ResNet-50, EfficientNet-v2, GoogLeNet, and Inception-v3 achieve a success rate of 97\%. However, AlexNet and ViT have the lowest success rates, at 94\% and 89\%, respectively.

    Execution time per attack also varies across models. EfficientNet-v2 and NiN demonstrate the fastest execution times, requiring approximately 0.31 seconds and 0.29 seconds per image, respectively. Inception-v3 also shows relatively quick execution at 1.14 seconds per image. In contrast, ViT incurs the highest computational overhead, with an average execution time of 2.36 seconds per image.
    
    Additionally, to get some generic experimental results, we run our method with several real-world datasets such as EuroSAT, GTSRB, and HAM10000 dataset. These results are also presented in Table~\ref{tab:mean_metrics}. For the EuroSAT dataset, ResNet-50 maintains high confidence (0.95) with minimal perturbation (0.02\%) and high SSIM (0.98), taking 26 iterations and 1.27 seconds per attack. For the GTSRB dataset, CNN\_GTSRB demonstrates high confidence (0.98), minimal perturbation (0.06\%), and moderate SSIM (0.82), however, has the longest execution time of 3.61 seconds. On the HAM10000 dataset, ResNet-50 achieves high confidence (0.97) with minimal perturbation (0.01\%) and high SSIM (0.99), requiring 15 iterations and 0.60 seconds per attack.
    \begin{table*}[!t]
        \begin{tabular*}{\textwidth}{@{\extracolsep{\fill}}ll*{6}{c}}
            \toprule
            Model & Dataset & Confidence$\uparrow$ & Perturbation & SSIM$\uparrow$ & Iterations & Success$\uparrow$ &   Time$\downarrow$\\
            \midrule
            ResNet-50* & ImageNet & 0.97 & 2.14\% & 0.99 & 29 & 0.97 & 0.37 s \\
            AlexNet* & ImageNet & 0.97 & 9.08\% & 0.92 & 25 & 0.94 & 0.52 s \\
            EfficientNet-v2** & ImageNet & 0.97 & 3.37\% & 0.98 & 33 & 0.97 & 0.31 s \\
            GoogLeNet** & ImageNet & 0.97 & 3.45\% & 0.97 & 33 & 0.97 & 1.48 s \\
            Inception-v3* & ImageNet & 0.97 & 2.35\% & 0.99 & 38 & 0.97 & 1.14 s \\
            ViT* & ImageNet & 0.96 & 11.27\% & 0.89 & 67 & 0.89 & 2.36 s \\
            LeNet-5*** & MNIST & 0.98 & 1.29\% & 0.63 & 8 & 0.95 & 0.29 s \\
            CNN (Defensive Distillation)*** & MNIST & 0.99 & 1.28\% & 0.64 & 10 & 0.96 & 0.12 s \\
            GoogLeNet*** & CIFAR-10 & 0.97 & 3.04\% & 0.96 & 16 & 0.94 & 1.38 s \\
            NiN***	& CIFAR-10	& 0.98 & 3.76\% &	0.95 &	12 & 1.00 & 0.29 s\\
            ResNet-20*** & CIFAR-10 & 0.98 & 1.66\% & 0.99 & 14 & 1.00 & 0.32 s\\
            CNN (Defensive Distillation)*** & CIFAR-10 & 0.99 & 6.91\% & 0.92 & 10 & 0.94 & 0.12 s\\
            ResNet-50*** & EuroSAT & 0.95 & 0.02\% & 0.98 & 26 & 0.92 & 1.27 s\\
            CNN\_GTSRB*** & GTSRB & 0.98 & 0.06\% & 0.82 & 11 & 0.99 & 3.61 s\\
            ResNet-50*** & HAM10000 & 0.97 & 0.01\% & 0.99 & 15 & 0.99 & 0.60 s\\
            \bottomrule
        \end{tabular*}
        \caption{The performance of ET Deepfool on various classifiers. These are the mean values from our experiment results. Here, * denotes the model is run on an RTX 3070 Ti machine. Besides, ** denotes the RTX 3060 Ti machine, and *** denotes the GTX 1050 Ti machine used to run the classifier. Also, the time column presents the average time it takes to run on a single image. Moreover, $\uparrow$ indicates the higher the score, the better results, and vice versa} 
        \label{tab:mean_metrics}
    \end{table*}

    \begin{figure*}[!t]
        \centering
        \includegraphics[width=0.8\linewidth]{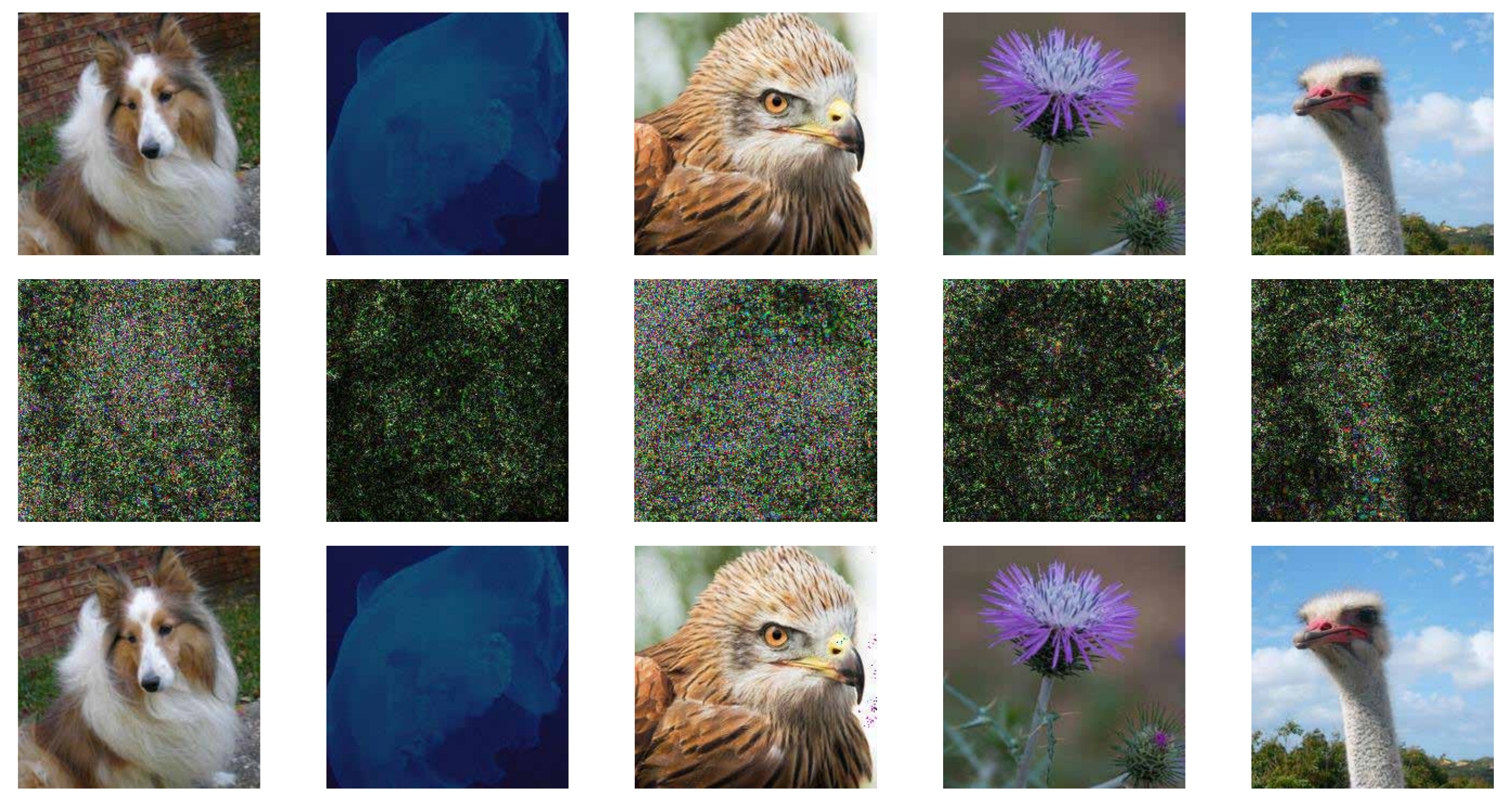}
        \caption{The 5 randomly chosen images, from the ImageNet dataset, shown in Table~\ref{tab:single_image_perturbations}, and their respective adversarial examples are illustrated in three rows. The first row represents the original images. The perturbations are shown in the second row, scaled 20 times for better visibility. Finally, the third row presents the perturbed images}
        \label{fig:sample_adversarial_examples}
    \end{figure*}
    
    \begin{table*}[!t]
        \begin{tabular*}{\textwidth}{@{\extracolsep{\fill}}l*{6}{c}}
            \toprule
            & \multicolumn{6}{c}{Perturbation Amount}\\ 
            \cmidrule(l){2-7} 
            Image No.   & ResNet-50      & AlexNet       & EfficientNet      & GoogLeNet     & Inception-v3     & ViT\\ 
            \midrule
            3.jpg       & 2.14\%        & 8.40\%        & 3.34\%            & 4.33\%        & 5.54\%           & 6.43\% \\
            63.jpg      & 0.86\%        & 7.93\%        & 0.47\%            & 1.84\%        & 0.65\%           & 3.60\% \\
            328.jpg     & 2.84\%        & 9.67\%        & 6.51\%            & 5.86\%        & 2.51\%           & 25.95\% \\
            1125.jpg    & 1.14\%        & 7.11\%        & 2.09\%            & 1.87\%        & 1.64\%           & 1.75\% \\
            1398.jpg    & 1.25\%        & 5.29\%        & 2.32\%            & 2.28\%        & 1.76\%           & 15.43\% \\
            \bottomrule
        \end{tabular*}
        \caption{Percentage of perturbations needed, on five randomly chosen sample images from the ImageNet validation dataset}
        \label{tab:single_image_perturbations}
    \end{table*}

    \begin{table*}[!t]
    \centering
    \begin{tabular*}{\textwidth}{@{\extracolsep{\fill}}lcccccc@{}}
        \toprule
        Minimum Confidence & Success Rate & Perturbation & SSIM & Iterations & Time & Confidence \\ \midrule
        No tuning          & 0.99         & 1.44         & 0.99 & 23         & 2.07 & 0.35 \\
        0.50                 & 0.99         & 1.55         & 0.99 & 24         & 2.73 & 0.66 \\
        0.70                 & 0.99         & 1.73         & 0.99 & 26         & 2.40 & 0.81 \\
        0.90                 & 0.98         & 2.03         & 0.99 & 29         & 2.85 & 0.94 \\
        \bottomrule
    \end{tabular*}
    \caption{Impact of confidence hyperparameters on ET DeepFool's performance on ImageNet}
    \label{tab:confTuning_imagenet}
    \end{table*}

    \begin{table*}[!ht]
    \centering
    \begin{tabular*}{\textwidth}{@{\extracolsep{\fill}}lcccccc@{}}
        \toprule
        Minimum Confidence & Success Rate & Perturbation & SSIM & Iterations & Time & Confidence \\ \midrule
        No tuning          & 0.99         & 0.008        & 0.99 & 6          & 0.52 & 0.53 \\
        0.50                 & 0.98         & 0.008        & 0.98 & 8          & 0.76 & 0.57 \\
        0.70                 & 0.98         & 0.009        & 0.99 & 12         & 1.10 & 0.80 \\
        0.90                 & 0.97         & 0.010        & 0.99 & 14         & 1.28 & 0.95 \\
        \bottomrule
    \end{tabular*}
    \caption{Impact of confidence hyperparameters on ET DeepFool's performance on HAM10000}
    \label{tab:confTuning_ham}
    \end{table*}

    \begin{table*}[!ht]
    \centering
    \begin{tabular*}{\textwidth}{@{\extracolsep{\fill}}lcccccc@{}}
        \toprule
        Minimum Confidence & Success Rate & Perturbation & SSIM & Iterations & Time & Confidence \\ \midrule
        No tuning          & 0.98         & 0.026        & 0.98 & 19         & 1.02 & 0.53 \\
        0.50                 & 0.98         & 0.029        & 0.98 & 22         & 1.27 & 0.60 \\
        0.70                 & 0.98         & 0.033        & 0.97 & 25         & 1.43 & 0.79 \\
        0.90                 & 0.98         & 0.039        & 0.97 & 28         & 1.60 & 0.93 \\
        \bottomrule
    \end{tabular*}
    \caption{Impact of confidence hyperparameters on ET DeepFool's performance on EuroSAT}
    \label{tab:confTuning_eurosat}
    \end{table*}

    \begin{table*}[!ht]
    \centering
    \begin{tabular*}{\textwidth}{@{\extracolsep{\fill}}lcccccc@{}}
        \toprule
        Minimum Confidence & Success Rate & Perturbation & SSIM & Iterations & Time & Confidence \\ \midrule
        No tuning            & 1.00         & 0.042        & 0.83 & 6       & 0.15 & 0.65 \\
        0.50                 & 1.00         & 0.043        & 0.83 & 7       & 0.16 & 0.72 \\
        0.70                 & 1.00         & 0.046        & 0.82 & 9       & 0.18 & 0.85 \\
        0.90                 & 1.00         & 0.052        & 0.82 & 10      & 0.21 & 0.96 \\
        \bottomrule
    \end{tabular*}
    \caption{Impact of confidence hyperparameters on ET DeepFool's performance on GTSRB}
    \label{tab:confTuning_gtsrb}
    \end{table*}


    
    To show some specific results, we choose five random images from the ImageNet validation dataset. The images are shown in Figure ~\ref{fig:sample_adversarial_examples}. We perform our method with these five images for various models and present the result in Table~\ref{tab:single_image_perturbations}. Here, We can see that after running our method, the successful samples require only a small amount of perturbations while maintaining a high confidence score.
    Overall, we find our method to be effective in various degrees for most of these classifiers. The results provide insights into the comparative strengths and weaknesses of the given image classifiers under adversarial conditions, which can aid in developing improved defense mechanisms and enhancing the overall robustness of image classifiers.
    
    We investigate the impact of varying confidence hyperparameters on the performance of the ET DeepFool algorithm across four distinct datasets: ImageNet, HAM10000, EuroSAT, and GTSRB. The results are presented in Tables~\ref{tab:confTuning_imagenet} -~\ref{tab:confTuning_gtsrb}.

    For the ImageNet dataset, the algorithm demonstrated a consistently high success rate across all confidence levels, ranging from 0.98 to 0.99. Perturbation values increased from 1.44 to 2.03 as the confidence score increased from 0.35 to 0.94. SSIM remained stable at 0.99, indicating minimal impact on image similarity. However, the number of iterations and computation time also increased with higher confidence, from 23 iterations (2.07 seconds) to 29 iterations (2.85 seconds) per image.

    Similar trends were observed with the HAM10000 dataset. The success rate exhibited a slight decrease from 0.99 to 0.97 as the confidence parameter increased. Perturbation values showed a marginal increase from 0.008 to 0.010. The SSIM remained mostly constant between 0.98 and 0.99. The number of iterations required grew from 6 to 14, with corresponding increases in time from 0.52 seconds to 1.28 seconds. The confidence values ranged from 0.53 to 0.95.

    For the EuroSAT dataset, the success rate remained constant at 0.98 regardless of the confidence parameter. Perturbation values increased from 0.026 to 0.039, and SSIM experienced a slight decrease from 0.98 to 0.97. The number of iterations increased from 19 to 28, with computation times increasing from 1.02 seconds to 1.60 seconds. Confidence values ranged from 0.53 to 0.93.

    The GTSRB dataset showed a consistent success rate of 1.00 across all confidence levels. Perturbation values increased from 0.042 to 0.052, while SSIM decreased slightly from 0.83 to 0.82. The number of iterations rose from 6.32 to 9.80, and the time required increased from 0.15 seconds to 0.21 seconds. The confidence values ranged from 0.65 to 0.96.

    Across all datasets, higher confidence hyper-parameters resulted in increased perturbation and iterations, leading to longer computation times. Success rates remained high and relatively stable, showcasing the robustness of the ET DeepFool algorithm. SSIM values were mostly stable, with minor reductions at higher confidence levels, indicating slight reductions in image similarity. These results highlight the trade-offs associated with achieving higher confidence in perturbations. While higher confidence levels lead to more significant perturbations and increased computational efforts, the success rates and SSIM values remain robust, demonstrating the effectiveness of ET DeepFool across different datasets.

        \begin{figure*}[!t]
        \centering
        \includegraphics[width=0.8\linewidth]{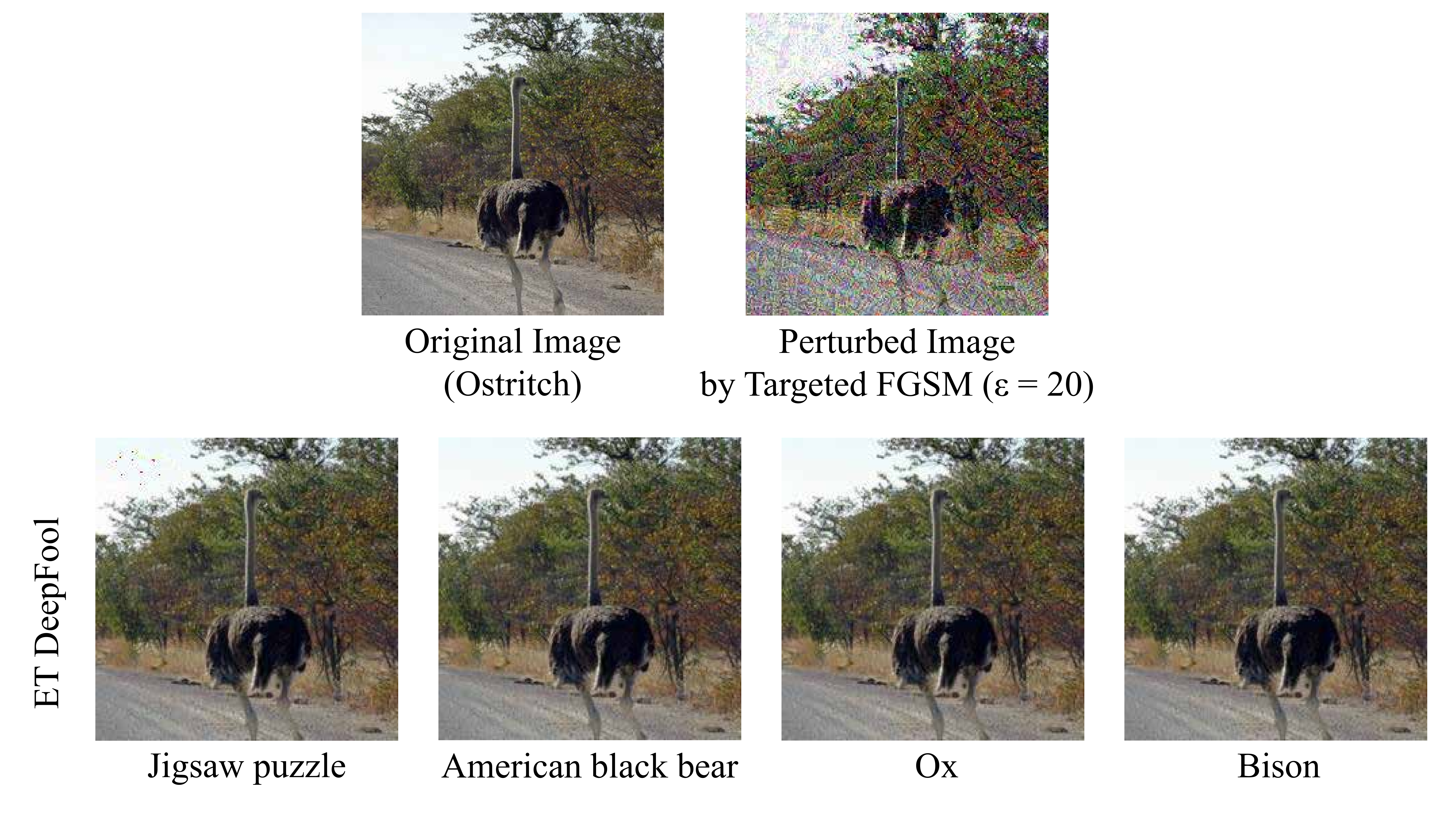}
        \caption{Visual comparison with Targeted FGSM using an image, original image from the ImageNet dataset}
        \label{fig:tfgsm_vs_etdeepfool}
    \end{figure*}

\section{Discussion}
\label{discussion}

    During the development of the ET DeepFool method, we observe an interesting phenomenon. While the original method exhibits the ability to misclassify an image with a minimal amount of perturbation, we notice that the confidence score associated with the perturbed image is often very low, shown in Figure~\ref{fig:DeepFool_vs_TargetedDeepFool}. 

    To tackle this issue, we introduce a confidence threshold as a specific hyperparameter that allows us to specify a minimum confidence level to be attained. We aim to enhance the overall confidence of the perturbed images while maintaining their effectiveness in inducing misclassification to specific classes by incorporating this hyperparameter into our Enhanced Targeted DeepFool approach which is not an option for the original DeepFool method as well as other existing works. One consequence of introducing this hyperparameter is an increase in the number of perturbations added to the original image. However, based on our experiments, we find that the additional perturbations are negligible in magnitude.

    In Tables~\ref{tab:quantitative_comparison_ImageNet},~\ref{tab:quantitative_comparison_MNIST}, and~\ref{tab:quantitative_comparison_CIFAR10}, we demonstrate that our method consistently outperforms other attacks. To ensure fair comparisons, we evaluate our method under the same conditions as the other reported methods. Additionally, in Table~\ref{tab:quantitative_comparison_ImageNet}, we include results from a more extensive evaluation using 50,000 samples across 1,000 classes for a comprehensive comparison. Targeted FGSM and C\&W show a surprising 100\% success rate. However, previous works have not examined the confidence with which successful attacks can induce misclassification of an image while maintaining minimal perturbations. On the other hand, in Table 1 of their paper, the authors of Targeted FGSM have stated that their attack achieves a 100\% success rate only when using the ResNet-50 model, in addition with a heavy amount of distortion (See Figure~\ref{fig:tfgsm_vs_etdeepfool} and Table~\ref{tab:comparison_tfgsm}). In Inception-v3 they have a success rate of 0.66 whereas we have a success rate of 0.97. Our attack has been executed on the sample image provided, and the results depicted in Figure~\ref{fig:tfgsm_vs_etdeepfool} reveal that our approach generates perturbed images that closely resemble the original, unlike the outcomes obtained through Targeted FGSM. Moreover, all of the existing works focus solely on the success rate, without considering the structural similarity of the images as shown in Table~\ref{tab:qualitative_comparison}. However, we make sure wherever it gets successful, it does that with the highest confidence and lowest perturbation possible.

    \begin{table*}[!t]
        \centering
        \begin{tabular*}{\textwidth}{@{\extracolsep{\fill}}lccclcc@{}}
            \toprule
            Attack Name                             & Success       & Confidence    & Perturbation  & Network   & Sample Size    & Classes \\ \midrule
            Targeted FGSM~\cite{tfgsm}              & 1.00         & -             & -             & ResNet-50     & 1000       & - \\
            Targeted FGSM              & 0.66         & -             & -             & Inception-v3     & 1000       & - \\
            C\&W~\cite{carlini2017}                 & 1.00         & -             & -             & Inception-v3  & 1000       & 100 \\
            DT UAP~\cite{dt_uap}                    & 0.73       & -             & -             & ResNet-50     & -          & - \\
            Po+Trip~\cite{PoTrip}                   & 0.85       & -             & -             & ResNet-50     & 1000       & - \\
            \textbf{ET DeepFool (Proposed) }                 & \textbf{0.98}       & \textbf{0.98*}      & \textbf{2.12\%}        & \textbf{ResNet-50}     & \textbf{1000}      & \textbf{1000} \\ 
            \textbf{ET DeepFool (Proposed) }                & \textbf{0.98}       & \textbf{0.97*}      & \textbf{2.53\%}        & \textbf{Inception-v3}     & \textbf{1000}      & \textbf{100} \\ 
            \textbf{ET DeepFool (Proposed) }                & \textbf{0.97}       & \textbf{0.97*}      & \textbf{2.14\%}        & \textbf{ResNet-50}     & \textbf{50000}      & \textbf{1000} \\ 
            \textbf{ET DeepFool (Proposed) }                & \textbf{0.97}       & \textbf{0.97*}      & \textbf{2.35\%}        & \textbf{Inception-v3}     & \textbf{50000}      & \textbf{1000} \\ 
            \bottomrule
        \end{tabular*}
        \caption{Comparison among existing attacks and our proposed attack on the ImageNet dataset. Reported values are sourced from their respective papers. Fields marked with '-' were not reported in the original studies. '*' represents a variable that can range from 0 to 100, depending on our preference} 
        \label{tab:quantitative_comparison_ImageNet}
    \end{table*}

    \begin{table*}[!t]
        \centering
        \begin{tabular*}{\textwidth}{@{\extracolsep{\fill}}lcccl@{}}
            \toprule
            Attack Name                             & Success       & Confidence    & Perturbation  & Network \\ \midrule
            C\&W~\cite{carlini2017}                 & 1.00         & -             & -             & CNN (Defensive Distillation) \\
            JSMA~\cite{jsma}                        & 0.97       & -             & 4.03\%        & Custom \\
            Targeted DeepFool~\cite{tardeep}        & 0.97       & -             & 2.28\%        & CNN (Defensive Distillation) \\
            \textbf{ET DeepFool (Proposed) }            & \textbf{0.96}       & \textbf{0.99*}      &\textbf{1.28\%}         & \textbf{CNN (Defensive Distillation)} \\
            \bottomrule
        \end{tabular*}
        \caption{Comparison among existing attacks and our proposed attack on the MNIST dataset. Reported values are sourced from their respective papers. Fields marked with '-' were not reported in the original studies. '*' represents a variable that can range from 0 to 100, depending on our preference} 
        \label{tab:quantitative_comparison_MNIST}
    \end{table*}

    \begin{table*}[!t]
        \centering
        \begin{tabular*}{\textwidth}{@{\extracolsep{\fill}}lcccl@{}}
            \toprule
            Attack Name                         & Success       & Confidence    & Perturbation  & Network \\ \midrule
            C\&W~\cite{carlini2017}             & 1.00         & -             & -             & CNN (Defensive Distillation) \\
            DT UAP~\cite{dt_uap}                & 0.81       & -             & -             & ResNet-20 \\
            One Pixel~\cite{Su_2019}            & 0.23       & 0.75          & -             & Network in Network~\cite{nin} \\
            Targeted DeepFool~\cite{tardeep}    & 0.77       & -             & 11.00\%          & CNN (Defensive Distillation) \\
            \textbf{ET DeepFool (Proposed) }             & \textbf{1.00}       & \textbf{0.98*}          & \textbf{1.66\%}        & \textbf{ResNet-20} \\
            \textbf{ET DeepFool (Proposed) }             & \textbf{0.94}       & \textbf{0.99*}          & \textbf{6.91\%}       & \textbf{CNN (Defensive Distillation)} \\
            \textbf{ET DeepFool (Proposed) }             & \textbf{1.00}       & \textbf{0.98*}          & \textbf{3.76\%}        & \textbf{Network in Network} \\
            \bottomrule
        \end{tabular*}
        \caption{Comparison among existing attacks and our proposed attack on the CIFAR-10 dataset. Reported values are sourced from their respective papers. Fields marked with '-' were not reported in the original studies. '*' represents a variable that can range from 0 to 100, depending on our preference} 
        \label{tab:quantitative_comparison_CIFAR10}
    \end{table*}

    \begin{table*}[!t]
        \begin{tabular*}{\textwidth}{@{\extracolsep{\fill}}ll*{4}{c}}
            \toprule
            & \multicolumn{3}{c}{Confidence}\\ 
            \cmidrule(l){3-5} 
            Perturbed Label      & Network       & Non-Targeted FGSM    & Targeted FGSM    & ET DeepFool (Proposed)\\ 
            \midrule
            Jigsaw puzzle        & ResNet-50     & -                    & 0.35             & 0.98\\
            American black bear  & ResNet-50     & 0.20                 & -                & 0.96\\
            Ox                   & Inception-v3  & -                    & 0.35             & 0.98\\
            Bison                & Inception-v3  & 0.45                 & -                & 0.99\\
            \bottomrule
        \end{tabular*}
        \caption{Comparison of confidence score between Targeted FGSM and ET DeepFool}
        \label{tab:comparison_tfgsm}
    \end{table*}

    \begin{sidewaystable}
        \centering
        \resizebox{\textwidth}{!}{
        \begin{tabular}{@{\extracolsep{\fill}}llclcc@{}}
            \hline
Attack Name & Dataset & Classes & Network Architectures & \begin{tabular}[c]{@{}c@{}}Confidence \\ Tuning\end{tabular} & \begin{tabular}[c]{@{}c@{}}SSIM \\ metric\end{tabular} \\ \hline
Targeted FGSM~\cite{tfgsm} & ImageNet & 1000 & \begin{tabular}[c]{@{}l@{}}ResNet-50, VGG-16, VGG-19, Inception-v3, \\ Xception, Inception-ResNet-v2\end{tabular} & - & - \\
C\&W~\cite{carlini2017} & MNIST, CIFAR-10, ImageNet & 100 & CNN with Defensive Distillation, Inception-v3 & - & - \\
JSMA~\cite{jsma} & MNIST & 10 & DNN based on LeNet Architecture & - & - \\
DT UAP~\cite{dt_uap} & \begin{tabular}[c]{@{}l@{}}CIFAR-10, GTSRB, EuroSAT, \\ YCB, ImageNet\end{tabular} & 1000 & VGG-16, ResNet-20, Inception-v3, ResNet-50, MobileNet-v2 & - & - \\
One Pixel~\cite{Su_2019} & MNIST, CIFAR-10, ImageNet & 1000 & All Convolutional Network, Network in Network, VGG & - & - \\
Po+Trip~\cite{PoTrip} & ImageNet & 1000 & \begin{tabular}[c]{@{}l@{}}Inception-v3, Inception-v4, Inception-ResNet-v2, \\ Resnet-v2-\{50,101,152\}, ens3-advInception-v3, \\ ens4-adv-Inception-v3, ens-adv-Inception-ResNet-v2\end{tabular} & - & - \\
Targeted DeepFool~\cite{tardeep} & MNIST, CIFAR-10 & 10 & CNN with Defensive Distillation & - & - \\ 
\begin{tabular}[c]{@{}l@{}}\textbf{Enhanced Targeted} \\ \textbf{DeepFool (Ours)}\end{tabular} & \begin{tabular}[c]{@{}l@{}}\textbf{MNIST, CIFAR-10, ImageNet,} \\ \textbf{EuroSAT, HAM10000, GTSRB}\end{tabular} & \textbf{1000} & \begin{tabular}[c]{@{}l@{}}\textbf{LeNet-5, ResNet-20, ResNet-50, AlexNet, EfficientNet-v2, Network in Network} \\ \textbf{CNN with Defensive Distillation, GoogLeNet, Inception-v3, VisionTransformer}\end{tabular} & $\checkmark$ & $\checkmark$ \\ \hline
        \end{tabular}
        }
        \caption{Qualitative comparison of different image classification attack methods with our proposed one. "Classes" refers to the number of targeted misclassification categories} 
        \label{tab:qualitative_comparison}
    \end{sidewaystable}

    \begin{table*}[ht]
        \begin{tabular*}{\textwidth}{@{\extracolsep{\fill}}lll}
            \toprule
            Algorithm & Complexity \\
            \midrule
            Original DeepFool & $O(NC)$ \\
            Basic Targeted DeepFool & $O(N)$ \\
            Recursive Targeted DeepFool & $O(RN)$ \\
            Enhanced Targeted DeepFool (Proposed) & $O(N)$ \\
            \bottomrule
        \end{tabular*}
        \caption{Time complexity analysis of various DeepFool algorithms. Here, $C$ represents the number of classes, $N$ is the maximum number of iterations, and $R$ is the maximum recursion depth}
        \label{tab:deepfool_complexity}
    \end{table*}

    \begin{figure*}[!ht]
        \centering
        \includegraphics[width=0.8\linewidth]{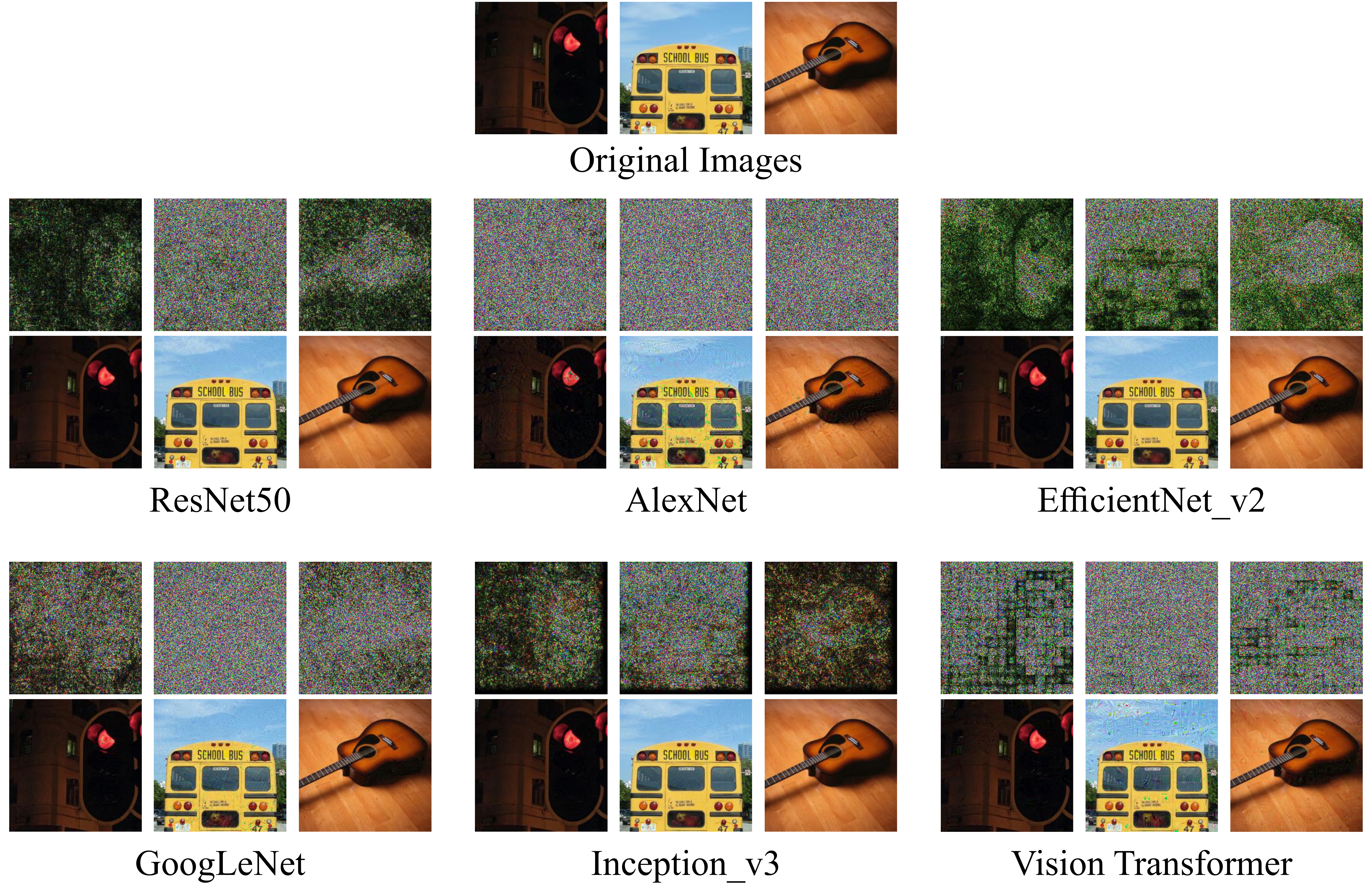}
        \caption{A few sample images of the ImageNet dataset from our experiments. The perturbed classes are as follows: Traffic light as Manhole cover, School bus as Ambulance, Acoustic guitar as Assault Rifle. Perturbations are shown in the first row, scaled 20 times for better visibility}
        \label{fig:samples}
    \end{figure*}

    
    The Targeted DeepFool algorithm by Gajjar et al., uses the MNIST dataset, which is a $28\times28$ pixel grayscale dataset of numbers, and the CIFAR-10 dataset which only contains 10 classes~\cite {tardeep}. The images in these datasets are not suitable to compare against real-life images. Their approach has a 97.83\% success rate which is a bit higher than our 96.65\%. However, our approach generates images that have a lower perturbation rate (1.29\%) as shown in Table~\ref{tab:quantitative_comparison_MNIST}. In Table~\ref{tab:quantitative_comparison_CIFAR10}, we can see that they get a 77\% success rate for CIFAR-10 whereas our approach gets a 94.22\% success rate with significantly less perturbation rate (3.04\%) than their 11\%. They also claim that the adversarial images 
    are visually indistinguishable from the original ones, however, this claim is questionable, as the images are too small to assess the level of distortion. Furthermore, although their basic iterative approach shares the same time complexity as ours, it under performs compared to their most efficient method, the recursive algorithm, which is more time-consuming, as demonstrated in Table~\ref{tab:deepfool_complexity}.
    Additionally, in contrast to the majority of untargeted attacks in the literature, our study does not seek to elucidate the degradation in model performance when retrained with perturbed images; consequently, we have chosen not to incorporate this aspect in our analysis.
    


    In our results, we observe that ViT performs better against our attack. Benz et al., also support this claim as their results show that ViT has a better success rate against vanilla DeepFool~\cite{benz2021adversarial}. One possible reason for this robustness is the unique architecture of the model. The process of splitting the image into multiple patches allows each patch to be processed individually, resulting in perturbations being distributed across the patches, as seen in Figure~\ref{fig:samples}. However, the robustness of ViT may not solely arise from patch splitting. The self-attention mechanism, which dynamically recalibrates the importance of each patch based on global context, likely plays a significant role in limiting the effectiveness of perturbations. Additionally, the inclusion of positional encoding helps the model retain spatial information, which may also contribute to the resilience against adversarial attacks. This combination of factors—patch-wise processing, attention, and positional encoding—could explain the comparatively higher mean perturbations observed in Table~\ref{tab:mean_metrics}, as perturbations in ViT are diffused across the global structure of the image, reducing the attack's overall impact.

    Our attack has far-ranging implications, as this can tell us what kind of image inputs are most likely to fail for a given model. Our current implementation uses only one target class, but this attack can be extended to include multiple classes with individual confidence levels, which can then expose which classes of images are more likely to be mistaken in an image recognition model. 

\section{Conclusion and Future Work}
\label{Conclusion}

    In this paper, we propose an algorithm, Enhanced Targeted DeepFool, which improves on the original DeepFool algorithm and makes it able to misclassify an image to a specific target class and achieve the desired amount of confidence needed to fool a classifier. We show that our algorithm is simple and requires less time complexity. We demonstrate that the algorithm performs well against various state-of-the-art image classifiers. Evidence is also presented on how our approach performs better than the existing one. We are very hopeful that the existing classifiers will become more robust to future attacks by training and fine-tuning the images generated by the proposed algorithm. 
    
    Our future plans include the extension of this attack for multiple classes, each with its own confidence level. This would make our method effectively capable of finding which classes of images are easier to mistake for another class, leading to a measurement of the robustness of an image recognition model given a specific sample of images. To the best of our knowledge, our work is the first perturbation procedure where a post-perturbation performance metric like the confidence level can be tuned to its arbitrary precision level. This allows us to have a look at how little perturbation, i.e. the minimal level of change, it requires to make any model mistake a class of images as another with strong error rate and misplaced confidence. Another area of potential future research lies in devising an approach that minimizes computational requirements. Although the proposed algorithm already demonstrates improved time complexity, more investigations can be done to optimize its computational demands to ensure broader practical applicability. Furthermore, our plan includes exploring how this attack would perform in a grey box and black box setting. Moreover, why the attack works less well for ViT and AlexNet can be further explored, so that a perturbation attack specifically tailored for them can be constructed. We hope that these findings contribute to the advancement of adversarial machine learning and provide a foundation for further exploration and refinement of targeted attack methods.

\section{Acknowledgment}

We gratefully acknowledge the computational resources provided by BRAC University and also the University of Alabama at Birmingham IT-Research Computing group for high-performance computing (HPC) support and CPU time on the Cheaha compute cluster, which was essential for the completion of this research.

\section*{Data Availability}
The data used in this research are publicly available from \url{https://www.image-net.org}, \url{https://www.cs.toronto.edu/~kriz/cifar.html}, \url{https://benchmark.ini.rub.de/}, \url{https://dataverse.harvard.edu/dataset.xhtml?persistentId=doi:10.7910/DVN/DBW86T}, \url{https://github.com/phelber/EuroSAT}, and \url{https://yann.lecun.com/exdb/mnist}. All data sources are appropriately cited in the references section. No new data were generated or analyzed in this study. Despite this, if required, the corresponding author will provide the data upon request.

\bibliography{Bibliography}

\end{document}